\documentclass[acmsmall]{acmart}

\setcopyright{none}

\usepackage{subcaption} 
\usepackage[utf8]{inputenc}
\usepackage{booktabs}
\usepackage{longtable}
\usepackage{tabularx,array}

\renewcommand\footnotetextcopyrightpermission[1]{} 
\setcopyright{none} 
\acmVolume{}
\acmNumber{}
\acmArticle{}
\begin{document}

\title{I don’t Want You to Die: A Shared Responsibility Framework for Safeguarding Child-Robot Companionship}


\author{Fan Yang}
\authornote{Corresponding Author}
\email{trovato@corporation.com}
\affiliation{%
  \institution{University of South Carolina}
  \city{Dublin}
  \state{Ohio}
  \country{USA}
}

\author{Renkai Ma}
\affiliation{%
 \institution{University of Cincinnati}
 \city{Cincinnati}
 \state{Ohio}
 \country{USA}}
 \email{renkai.ma@uc.edu}

\author{Yaxin Hu}
\affiliation{%
  \institution{University of Wisconsin--Madison}
  \city{Madison}
  \country{USA}}
\email{yaxin.hu@wisc.edu}

\author{Michael D. Rodgers}
\affiliation{%
  \institution{University of South Carolina}
  \city{Columbia}
  \country{USA}}
\email{mr182@email.sc.edu}

\author{Lingyao Li}
\affiliation{%
  \institution{University of South Florida}
  \city{Tampa}
  \country{USA}}
\email{lingyaol@usf.edu}

\renewcommand{\shortauthors}{Yang et al.}

\begin{abstract}

  Social robots like Moxie are designed to form strong emotional bonds with children, but their abrupt discontinuation can cause significant struggles and distress to children. When these services end, the resulting harm raises complex questions of who bears responsibility when children's emotional bonds are broken. Using the Moxie shutdown as a case study through a qualitative survey of 72 U.S. participants, our findings show that the responsibility is viewed as a shared duty across the robot company, parents, developers, and government. However, these attributions varied by political ideology and parental status of whether they have children. Participants' perceptions of whether the robot service should continue are highly polarized; supporters propose technical, financial, and governmental pathways for continuity, while opponents cite business realities and risks of unhealthy emotional dependency. Ultimately, this research contributes an empirically grounded shared responsibility framework for safeguarding child-robot companionship by detailing how accountability is distributed and contested, informing concrete design and policy implications to mitigate the emotional harm of robot discontinuation.

\end{abstract}




\maketitle

\section{INTRODUCTION}
Humans can form strong emotional bonds with companion technologies, especially artificial intelligence (AI) and robotic companions, through emotional responsiveness and personalized interaction ~\cite{adam2025supportive}. Users have been shown to develop deep emotional dependence and experience genuine grief when losing access to AI chatbots like Replika ~\cite{laestadius2024too, ta2020user, banks2024deletion}. This bond is even more profound with embodied, physical robots, whose social presence fosters deeper, more sustained emotional connections through multi-sensory engagement and trust-building ~\cite{lee2006physically, li2015benefit, wainer2006role, kiesler2008anthropomorphic}. 

These companion robots are particularly appealing to vulnerable populations, especially children. For example, 
Research shows children, particularly those with neurological disorders, are highly likely to anthropomorphize their robot companions and form strong bonds with them ~\cite{rovstvsinskaja2025unlocking, holeva2024effectiveness}, consistently describing them as best friends and family members ~\cite{zhao2025robot, constantinescu2022children}. Children are drawn to social robots for the robots' ability to engage and motivate interaction, their reduced social complexity relative to human agents, and the enhanced predictability of their behavioral responses ~\cite{dubois2024people, hou2022young}. 

\begin{figure*}[htbp] 
    \centering
    \begin{subfigure}[b]{0.28\textwidth}
        \centering
        \includegraphics[width=\textwidth]{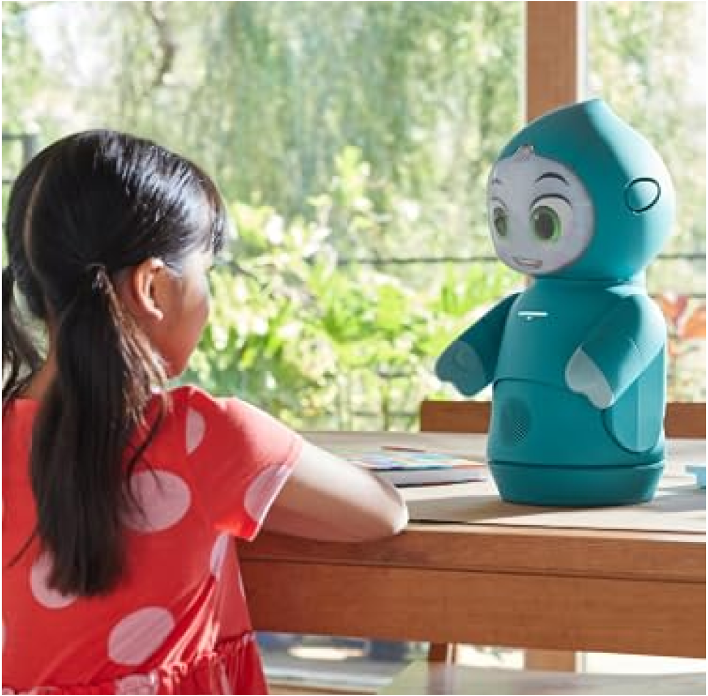} 
        \caption{A child interacting with the Moxie robot on a desk, highlighting its role as an educational companion.}
    \end{subfigure}
    \hfill 
    \begin{subfigure}[b]{0.26\textwidth}
        \centering
        \includegraphics[width=\textwidth]{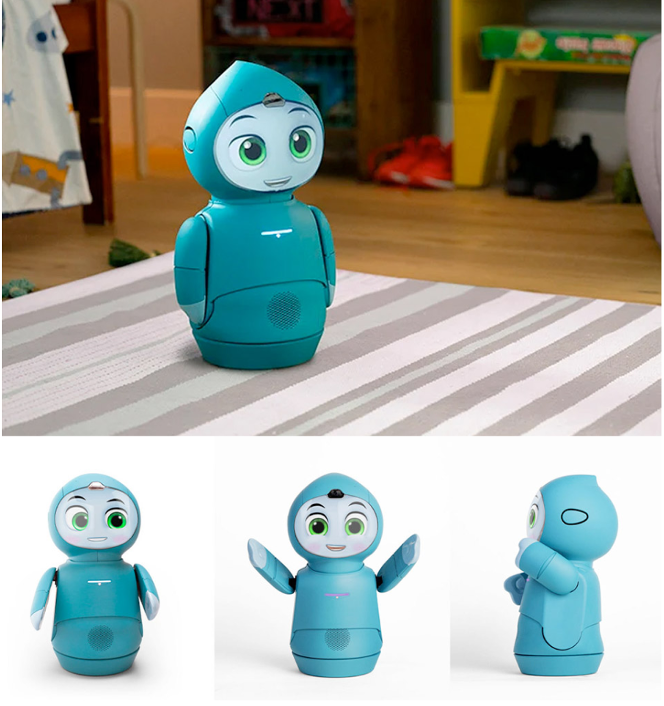} 
        \caption{Various views of the Moxie robot with its design and features from different angles.}
    \end{subfigure}
    \hfill
    \begin{subfigure}[b]{0.33\textwidth}
        \centering
        \includegraphics[width=\textwidth]{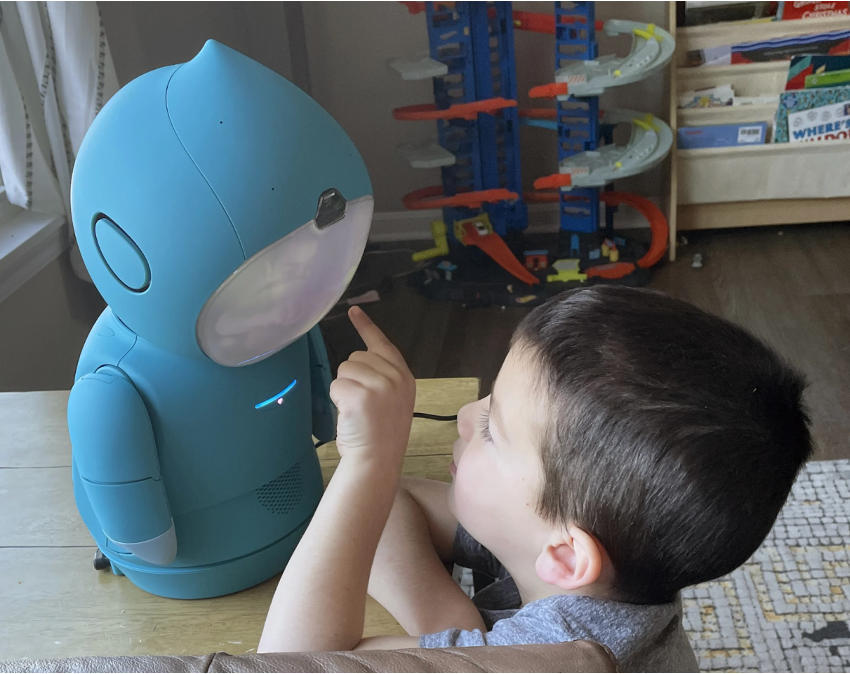} 
        \caption{A child engaging with the Moxie robot in diverse home environments.}
    \end{subfigure}
    \caption{The Moxie companion robot in different interaction scenarios and showcasing its design.}
    \label{fig:moxie_overview}
\end{figure*}

Consequently, the market for social robots designed specifically for childhood companionship and developmental support has expanded rapidly in recent years ~\cite{dawe2019can}. Among these emerging products, the award-winning Moxie robot stands out as a prominent example ~\cite{montalvo2024user}. Developed by Embodied Inc., Moxie was designed and marketed to facilitate children's acquisition of fundamental social-emotional competencies through play-based interactions, targeting skills such as turn-taking, eye contact, emotion regulation, and empathy ~\cite{cano2021affective}. Standing at approximately one foot tall with a soft-touch blue body, Moxie was intentionally designed with large, expressive green eyes and a round, teardrop-shaped head that creates a non-threatening, approachable appearance. In addition, Moxie was also given an 8ish-year-old boy's voice, making it especially easy to be identified by children as a friend rather than an authority figure (see Figure \ref{fig:moxie_overview}). Also, marketing strategies for such children's companion robots uniformly target parental anxieties by promising ``24/7'' support with phrases like ``always patient, always present'' ~\cite{gunawan2025promises}. These campaigns typically minimize the term ``robot,'' instead personifying the devices as constant ``companions'' or ``friends'' for children. Consequently, prominent examples like Moxie have engaged in over 4 million conversations since 2020, reaching tens of thousands of U.S. families ~\cite{MoxieRobotOurStory}.

Unsurprisingly, when companion technologies fail, the emotional fallout is devastating ~\cite{carter2020death}. This phenomenon has been observed in adults grieving the loss of AI companions like Replika ~\cite{de2024lessons, namvarpour2024uncovering}, but the emotional trauma is profoundly more severe for children, leaving parents at a loss in dealing with their grief ~\cite{gunawan2025promises}. When Embodied Inc. abruptly shut down its Moxie robot, social media videos of children sobbing and pleading to ``save'' their friend vividly illustrated this distress. This burgeoning companion-robot industry, however, operates in an alarming regulatory vacuum, with its products falling through the cracks of laws that have not kept pace with technology ~\cite{de2025unregulated}. Because no single entity currently bears clear responsibility, the complexity of the Moxie robot ecosystem, involving manufacturers, developers, parents, and regulators, demands a comprehensive examination of a shared-responsibility framework to safeguard children's emotional well-being ~\cite{mclennan2014exposing}.

Building on existing literature on shared responsibility in healthcare ~\cite{elwyn2012shared, bruch2024effects} and AI ethics ~\cite{arroyo2022shared, muhlhoff2025collective, lenk2017ethics}, our study employs the shutdown of the Moxie robot as a critical case to understand how such responsibility is attributed. To this end, we aim to investigate not only who the public holds accountable but also how these attributions differ across key demographics like parental status and political affiliation. Understanding these varied perspectives is important for developing an effective shared-responsibility framework and informing what further repair or coping is necessary for children and their families. Therefore, we ask two primary questions: 

\begin{itemize}
    \item[\textbf{RQ1:}] How is responsibility attributed for a child's suffering after losing the Moxie companion robot?
    \begin{itemize}
        \item[\textbf{RQ1a:}] Who are the key stakeholders identified as responsible, and what are their assigned responsibilities?
        \item[\textbf{RQ1b:}] How might these attributions vary across demographic variables, like parental status and political affiliation?
    \end{itemize}
    
    \item[\textbf{RQ2:}] How do people perceive the continuous service for the Moxie companion robot?
    \begin{itemize}
        \item[\textbf{RQ2a:}] Who are the key stakeholders responsible for this continuous service, and what are their responsibilities?
        \item[\textbf{RQ2b:}] Why do people oppose or express uncertainty about a continuous service?
    \end{itemize}
\end{itemize}

To answer these questions, we conducted a qualitative survey on $N=72$ U.S. participants. Through inductive thematic analysis on open-ended responses, we found that responsibility for a child's suffering was attributed to multiple stakeholders, including the robot company, parents, developers, marketers, and the government (RQ1a). These attributions and proposed solutions demonstrated divergence based on political affiliation (systemic corporate failure vs. individual parental judgment) and parental status (hands-on parenting vs. foundational critique) (RQ1b). Besides, perceptions of continuous service for the Moxie robot were highly polarized, with support nearly matching opposition (RQ2b), leading participants to propose detailed pathways for continuity, including technical mechanisms, sustainable financial models, and external governmental guarantees (RQ2a). These findings inform our discussion of a shared responsibility framework in child-robot companionship and its implications for design and policy.

Our paper makes three primary contributions to the fields of Human-Robot Interaction (HRI), AI Ethics, and Child-Robot Interaction. First, we provide a detailed empirical account of responsibility attribution surrounding the loss of a social companion robot, establishing a shared responsibility framework for mitigating emotional harm in children (see Figure \ref{fig:shared_responsibility_framework}) that extends accountability beyond the manufacturer to include parents, marketers, and regulators. Second, we reveal that the demographic variables, such as political and parental status, shape how the participants assign responsibility for the failures of a social robot, underscoring the challenge of establishing a uniform policy. Lastly, our study translates its findings into actionable design and policy recommendations for the ethical governance of embodied AI for children and their families.

\section{RELATED WORK}
\subsection{Robotic Bereavement}
Building on the concept of ``technological bereavement'' ~\cite{SustainabilityDirectoryBereavement}, the idea of robotic bereavement refers to affective, cognitive, and behavioral grief responses to the involuntary loss of access to a robotic companion with which a user has formed a significant emotional and relational bond. As various Internet of Things (IoTs) keep on permeating into our daily lives, technologies have now shifted from being a mere medium through which we process human loss to becoming the object of loss itself. We are now not only using AI-powered technologies to cope with death in real life (e.g., ~\cite{king2022exploring}), a phenomenon called ``thanatechnology'' ~\cite{ozdemir2021thanatechnology, sofka2020transition} or ``grieftech'' ~\cite{sri2025role, mahapatra2025ai}- but users are also increasingly facing the death of technologies themselves ~\cite{gunawan2025promises, wu2025silicon, zhao2025robot,carter2020death}.

Children's early stages of development render them at exceptionally high risks of robotic bereavement ~\cite{gunawan2025promises}, combined with their heightened susceptibility to anthropomorphic cues ~\cite{geerdts2016learning,festerling2022anthropomorphizing, van2020child}. Despite ongoing scholarly debating about children's ability to distinguish reality from fantasy before age 12 ~\cite{woolley2013revisiting}, recent research reveals a crucial nuance: although children may be more skeptical than traditionally suggested ~\cite{piaget2017child}, they show significantly higher belief rates when encountering novel entities in real-world contexts—precisely the scenario of robot companion interaction—compared to fantastical settings ~\cite{woolley2006effects}. This real-world bias becomes particularly pronounced when positive emotions such as happiness accompany the novel entity, with children's judgment becoming less critical regardless of the entity's actual nature.~\cite{carrick2006effects}. For companion robots like Moxie designed for children, this creates perfect conditions for belief formation: the physical presence of the robot signals ``real-world'' rather than artificiality, while their designed appearances and responsiveness trigger positive emotional associations that further suppress skeptical evaluation ~\cite{rogers2025designing}.

Research, albeit limited due to the novice stage of robot companions, has already warned about the devastating effects of robot ``death'' on children. When robots like Opportunity Rover, Jibo, and Kuri were discontinued, users flocked to social media to mourn their robot companions despite fully acknowledging their artificial nature ~\cite{carter2020death}. Scholars argued that the emotional attachment between humans and robots is genuine, sometimes equivalent to human-animal and human-human bonds ~\cite{wu2025silicon, kang2023robot}. As such, in a 4-year longitudinal study of 19 families who had educational reading robots, 18 families retained the robots despite the children aging out of the target demographic. This retention wasn't passive storage; families actively maintained these robots through charging rituals, display in meaningful locations, and integration into new household routines ~\cite{zhao2025robot}. This leads to a crucial ``right-to-use'' question when it comes to the shutdown of robot companions: should a robot companion be required to continue at least basic services that users, particularly children, have formed emotional dependencies on? And if so, who bears responsibility to ensure its fulfillment?

\subsection{Mitigating Risks in Human-Robot Bonds}
The practice regarding ethical and responsible artificial companions reveals a patchwork of reactive and unclear measures. On top of the over- and false promises regarding the nature and capabilities of an AI companion ~\cite{wu2025silicon}, users are at the mercy of companies' voluntary disclosure practices: Character.AI added hourly warnings only after a teen suicide, while Embodied issued a letter to parents that claimed to help them explain the shutdown of Moxie. These ``lip-service'' measures are fundamentally inadequate. Extensive research demonstrates that even non-anthropomorphic technologies generate powerful dependencies through habitual use—from video game addiction ~\cite{brechtelsbauer2025does} to problematic phone dependency ~\cite{drouin2015mobile}. The highly anthropomorphic design of companion robots amplifies these risks when repeatedly promoted to exploit users' vulnerabilities for social connections ~\cite{gunawan2025promises,adewale2025virtual}, deliberately fostering emotional and relational dependency beyond users' cognitive awareness ~\cite{wu2025silicon}. 


In the U.S., emerging state-level legislation in California and New York targets child protection in AI, but these measures primarily address chatbots, failing to account for the intensified attachment created by physical embodiment ~\cite{HuntonNYAICompanion, BellanCaliforniaAI}. Internationally, the regulatory landscape is inconsistent. The EU's AI Act classifies companion AI under transparency requirements rather than as high-risk, though actions like Italy's fine against Replika signal a growing awareness of emotional vulnerabilities ~\cite{EUAIACT, EDPBItaReplikaFine}. In contrast, Japan's AI Promotion Act prioritizes innovation with an emphasis on voluntary compliance ~\cite{BirdBirdJapanAI}, while global standards like UNESCO's AI Ethics Recommendation remain aspirational without enforcement mechanisms ~\cite{UNESCO_AI_Rec_2022}. This patchwork approach is being challenged by advocacy groups filing FTC complaints ~\cite{Chow2025ReplikaFTC} and watchdog organizations like Common Sense Media, which rate AI companions as "unacceptable" for minors ~\cite{CommonSense2025TalkTrust}.


These broad legislative gaps create specific policy failures for embodied companion robots. Critically, no jurisdiction has enacted right-to-repair legislation for these devices, leaving families helpless when a company fails and their child's "friend" becomes a brick ~\cite{boniface2024towards}. While data portability rights exist under regulations like GDPR Article 20 ~\cite{GDPR_Art20}, their technical implementation is underdeveloped. Industry-led solutions have also proven inadequate. Embodied Inc.'s OpenMoxie project, for instance, was presented as a remedy but its high technical barrier, last-minute notice, and lack of support reveal it as damage control rather than genuine consumer protection ~\cite{Beghtol2025OpenMoxie}. Ultimately, such "solutions" merely shift the responsibility for a crisis the company created onto families ill-equipped to manage it.

\subsection{Shared Responsibilities in Child-Robot Companionship}
Given the complexity of the emotional and relational bonds between children and robot companions as evidenced in the Moxie case and near absence of any regulatory guidelines to ensure protection of children when their robot companions face service shutdown, it is crucial to establish a shared responsibility framework ~\cite{stiggelbout2015shared} that identifies key stakeholders and holds them accountable to better protect children from suffering robotic bereavement.

Healthcare literature has long championed the theoretical framework of Shared Decision Making (SDM) ~\cite{elwyn2012shared, bruch2024effects} that distributes responsibility across patients, clinicians, and caregivers in treatment decisions, recognizing that optimal outcomes require coordinated input from all affected parties ~\cite{shay2015evidence}. Similarly, AI ethics scholarship is also increasingly advocating shared responsibility models ~\cite{arroyo2022shared, muhlhoff2025collective} that acknowledge how AI harm emerges from a chain of reactions between developers, deployers, marketers, and users rather than single points of failure.
Our study, therefore, uses the shared-responsibility approach as a conceptual lens, which moves beyond single-party blame to acknowledge collective actions and identify how different stakeholders like software developers ~\cite{rogers2025designing}, marketers ~\cite{gunawan2025promises}, regulators ~\cite{wu2025silicon}, parents ~\cite{rosman2025exploring, ho2025empowering} and more, can coordinate to prevent emotional harm to children.

\section{METHODS}

\subsection{Data Collection: Qualitative Survey and Participants}

Our study utilizes the Moxie shutdown as a case to empirically investigate shared responsibilities for robotic bereavement. To create a video stimulus for our qualitative survey, we systematically searched public social media platforms (TikTok, YouTube, Instagram) in August 2025 using the query: ``((``Moxie Robot'' AND shutdown) OR (``Moxie Robot'' AND goodbye) OR (``Moxie Robot'' AND farewell) OR (``Moxie Robot'' AND ``last conversation") OR (``Moxie Robot'' AND dying) OR (``Moxie Robot" AND dead) OR (``Embodied Moxie" AND goodbye) OR (``AI robot Moxie" AND shutdown))''. The search yielded 280 results, which were manually screened to 11 unique videos directly tied to the shutdown. From these, we selected four clips depicting children's distress and struggles and compiled them into a 4-minute video. This stimulus was then shown to N=72 participants on Prolific, following IRB approval.

We specifically employed a qualitative survey with open-ended questions (see Appendix \ref{appendix_surveydesign} for our survey design) to capture perceptions of responsibility attribution and service continuity. Participants were recruited through Prolific, a high-quality research platform ~\cite{palan2018prolific, douglas2023data}, with eligibility limited to U.S. residents aged 18 or older who are fluent in English. To ensure response diversity, we used quotas to achieve a balance in gender, political affiliation, and parental status (see Appendix \ref{appendix_demographic} for complete demographics). After providing consent via a Qualtrics link, participants watched the video of children reacting to the Moxie shutdown. They then responded to open-ended questions identifying who bore responsibility for the children's suffering and what measures could have prevented it. The survey concluded by probing their views on a "right to use"—whether services should be mandated to continue post-closure and who should be responsible for such measures.

\subsection{Data Analysis}
We performed an inductive thematic analysis ~\cite{Braun2006UsingPsychology} on the open-ended survey responses. One of the authors began the analysis by familiarizing themselves with the dataset, reading all responses to assess their depth and richness. Following this, the coder started assigning initial codes to the data in Google Sheets. For example, the statement, ``Make the robot not reliant on the internet or the company's servers for basic functions, so it can still work in some respect when the company dies,'' was assigned the initial code ``technical responsibility for offline functionality.'' After this initial coding, the author grouped related codes into sub-themes and consolidated these into the primary themes presented in our findings.

This process was conducted in two main stages. First, we analyzed the entire dataset to identify the overarching themes that address the shared responsibility framework (RQ1a) and the perceptions of continuous service (RQ2). Following this primary analysis, we conducted a secondary, stratified analysis to address RQ1b. For this stage, we filtered the dataset by participants' political affiliation and parental status, analyzing each subset inductively to identify the thematic patterns in how responsibility was attributed and how solutions were proposed. Throughout both stages, the research team discusses regularly to discuss the codes, themes, and associated quotations. These sessions were used to refine the thematic structure and resolve disagreements, which resulted in the thematic scheme used to structure our findings.

\section{RESULTS}
This section presents our findings on the attribution of responsibility for the Moxie robot's discontinuation and participants' perceived viability of its continuous service. Our findings first reveal a shared responsibility across stakeholders (Section~\ref{sec:shared_framework}) and then analyze how attributions of such responsibility or blame differ based on political ideology and parental status (Section~\ref{sec:political_divide} and Section~\ref{sec:parental_divide}) to answer RQ1. Subsequently, we explore the polarized views on this requirement, detailing both the practical pathways proposed to fulfill it (Section~\ref{sec:fulfilling_service}) and the key reasons for opposing it (Section~\ref{sec:opposing_service}) to answer RQ2.

\subsection{RQ1: Stakeholder Responsibility and Its Different Attributions}
In response to our RQ1, we found that participants attributed responsibility for a child’s suffering not to a single entity but to a shared responsibility of multiple stakeholders, and these perceived attributions varied based on participants' political ideology and parental status, whether they have children.

\subsubsection{A Shared Responsibility (RQ1a)}
\label{sec:shared_framework}
\mbox{}\

\begin{figure}[h]
    \centering
    \includegraphics[width=0.65\columnwidth]{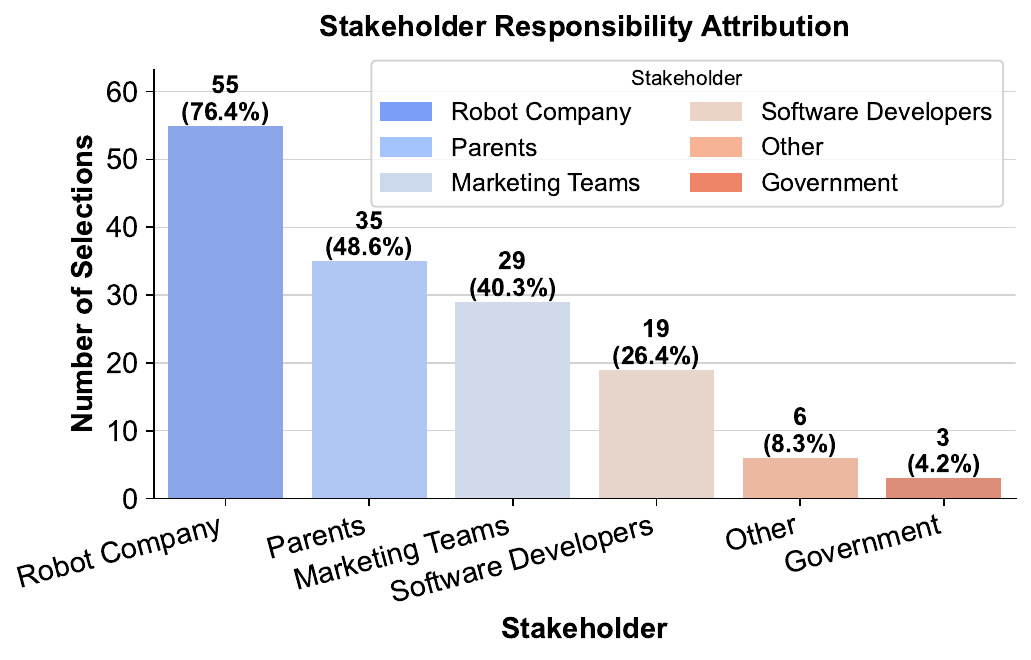}
    \caption{Multi-stakeholder responsibility attribution ($N=72$) in a child's suffering given Moxie robot discontinuation.}
    \label{fig:stakeholder_attribution}
\end{figure}

\noindent Participants did not attribute responsibility to a single entity but described a shared responsibility of the robot company, developers, marketers, parents, and government (see Figure~\ref{fig:stakeholder_attribution}), while the robot company was most frequently identified as a primary stakeholder. 

\textbf{Preventing the Robot Company’s Failure Through Sound Financial Management.}
Participants identified the foundational responsibility of maintaining a viable business to avoid bankruptcy, a duty shared by the Moxie company, its financiers, and government oversight. They first saw the company’s collapse as the primary cause of the children’s suffering, arguing it deployed a \textit{``failed business plan''} and was \textit{``not economically viable''}, as a participant stated:
\begin{quote}
\textit{It is the responsibility of a company to manage its financials properly and make sure they have enough money to continue on and not shut down at the first sign of distress.}
\end{quote}

This suggests that sound financial planning of the Moxie company is not just a business goal but an obligation when its product is designed to create emotional bonds. Participants also then extended this funding responsibility to stakeholders like financiers, stating the company failed because \textit{``unfortunately they ran out of money,''} while others suggested the government could have a role, proposing \textit{``programs to help keep failing companies afloat.''}

\textbf{Designing for Robot Product Longevity and End-of-Life.}
Participants argued that the Moxie robot company and its software developers/engineers are responsible for designing resilient robots that can outlive the company. The most common suggestion was to implement offline capabilities, as a participant stated: \textit{Make the robot not reliant on the internet or the company's servers for basic functions, so it can still work in some respect when the company dies.}

Another focused on the software, arguing that developers could easily \textit{``release a firmware that restores some basic functionality. It genuinely seems insane to me that this companion robot was coded to more or less simulate death... that was a choice.''} This technical foresight requires a clear contingency plan, as participants noted the company \textit{``didn't have a contingency plan for their servers.''} 

\textbf{Managing Emotional Attachment Through Shared Corporate and Parental Guidance.}
Participants assigned a dual duty to manage children's emotional attachment, holding the Moxie company responsible for ethical robot design and marketing while identifying parents as the ultimate guardians of their child's emotional well-being. First, participants argued that the company, marketing teams, and developers must be transparent and avoid predatory practices, stating they \textit{``designed the product in a way that would make kids attached.''} This responsibility extended to marketing/sales teams, who participants argued should not \textit{``advertise the product as a companion.''} To counter this, a participant called for transparent warnings, as commented: \textit{``The company should be required to clearly explain to parents that the children may develop deep emotional attachments to the robots, so the parents can make the best decision.''} This case frames transparency as a prerequisite for enabling parents to provide informed consent for purchasing and adopting the robot. 

Second, participants assigned the final responsibility for managing children's emotional attachment with the robot to parents. This begins with setting clear boundaries, with one participant stating it is \textit{``unhealthy to let your child form a relationship like that with a robot.''} One parent offered a clear directive: \textit{``Limit their childs time with moxie. And explain to their child that moxie is not a living thing.''} This places the onus on parents to frame the relationship in a healthy way. Participants also saw parents as responsible for teaching resilience and prioritizing real-world social skills, urging them to \textit{``set up actual play dates with real children friends''} and to use the discontinuation as a teachable moment for \textit{``explaining the emotions and difficulties of the human experience.''}

\subsubsection{Political Ideology Difference: Systemic Corporate Failure vs. Individual Parental Judgment (RQ1b)}
\label{sec:political_divide}
\mbox{}\

\noindent We further found an ideological divergence in how participants attribute responsibility and propose solutions. Democrats held the robot company accountable for systemic failures and advocated for systemic solutions (e.g., legislation). In contrast, Republicans centered on the personal responsibility of parents and proposed market-based solutions. Independents and Others articulated a balanced approach.

\textbf{Democrat: Holding the Company Accountable and Advocating for Systemic Solutions.}
Democrat participants focused on the systemic failures of the robot company and their software developers/engineers. A participant commented: \textit{``They were the ones that created the product and a failed business plan.''} Another framed the shutdown as an avoidable technical choice, stating:
\begin{quote}
\textit{The robot's functions should be independent of the company that created it. Doesn't it seem more logical to make a companion robot for children that is programmed to adapt without the company being involved?}
\end{quote}

This suggests the robot product’s failure was a preventable design flaw. This lens of corporate accountability extended directly to their proposed solutions: participants advocated for a hybrid model of corporate mandates and government intervention. A participant stated that the government could: \textit{``help these companies to get money with low interest to maintain their activities.''}

\textbf{Republican: Centering Parental Duty and Company Solutions.}
In contrast, Republicans centered their arguments on the responsibility of parents for permitting over-attachment and for failing to teach their children about resilience. A participant commented: \textit{``Because I felt that they depended on Moxie to parent their child. They should not have allowed their children to be so dependent.''} These participants also emphasized the parental duty to manage their children's expectations, as one stated: \textit{``Parents should teach their children that the robot is not real and things could happen that would make it not work the way it does now.''}

This focus on individual responsibility shaped their proposed solutions. Rather than calling for systemic oversight, they advocated for the robot company to achieve self-sufficiency through market-based discipline. One suggested a user-funded model where a \textit{``customer can pay 10 per month to keep the robot working.''} Republicans also viewed the company’s collapse as a normal market outcome. One explained: \textit{``The company doesn't have specific responsibilities other than trying to do a better job handling finances.''} This participant's comment suggests that business failure does not create a perpetual ethical duty to consumers.

\textbf{Independent/Other: Balancing Parent and Company Responsibility.}
Independents and Others shared a balanced view, assigning concurrent responsibilities to both parents and the robot company. They held parents responsible for moderating technology use, with one participant questioning the initial parental decision: \textit{``Why would you ever get your kid this thing? I mean, come on, what kind of parent are you?''} Meanwhile, they placed an obligation on the robot company to ensure a baseline of post-shutdown functionality. One participant offered a technical solution: \textit{``They should have made it so that the robot could just do pre-recorded feedback when they knew the company was going bankrupt.''}

\subsubsection{A Parental Status Difference: Hands-On Parenting vs. Critique on Foundational Decisions (RQ1b)}
\label{sec:parental_divide}
\mbox{}\

\textbf{Parents: Focusing on the Practice and Process of Parenting.}
Parental status also shaped participants' perspectives. Participants with children concentrated on the duties and processes of parenting with companion technologies, including the strategies to manage the child-robot relationship and guide the child's emotional development. One parent stated: \textit{``Robots or any form of technology should not be raising your kids. It should be used as a tool if you're going to use it, not a babysitter or part-time parent.''}

This focus on hands-on parenting extends to their proposed solutions, which are child-centered. They proposed support systems like a \textit{``managed phase-out program''} and access to a \textit{``counselor''} to help children process the transition. Another parent viewed the Moxie discontinuation as a pedagogical moment: \textit{``It's just like a pet. Loss is a part of life, and dealing with grief is a process. The parents should be helping their children deal with it by explaining the emotions and difficulties of the human experience.''} This perspective reframes the negative event as a valuable life lesson during parenting.

\textbf{Non-Parents: Critiquing the Foundational Decisions of the Robot Company and Parents.}
Participants without children tended to critique the foundational decisions made by both parents and the company, and condemned the parental choice to purchase the Moxie robot, viewing it as an inappropriate substitute for genuine companionship. One non-parent argued:
\begin{quote}
\textit{Socialize them with other children. Parent them and spend time with them... Let them be children. Or get them a dog!!!}
\end{quote}

This view suggests the problem was the initial parental choice to rely on technology for social development. Another saw the purchase as an error, stating, \textit{``They buy these, enabling the cycle.''} 

\subsection{RQ2: Pathways to Continuous Service of the Robot and the Reasons Against it}
\label{Sec:RQ2}
\begin{figure}[h]
    \centering
    \includegraphics[width=0.65\columnwidth]{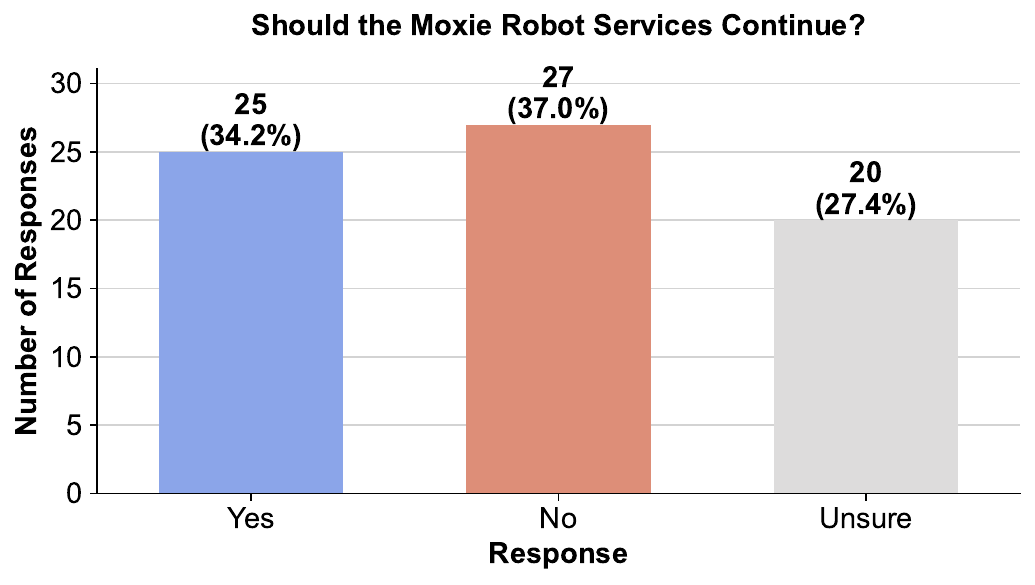} 
    \caption{Perception of continuous moxie service ($N=72$).}
    \label{fig:continuous_service_perception}
\end{figure}

In response to our RQ2, participants' perceptions of a continuous service of the Moxie robot were polarized, with support for and opposition against continuity nearly equal, and a substantial portion remaining uncertain, as shown in Figure~\ref{fig:continuous_service_perception}. Those who supported it identified a clear set of stakeholders, primarily the robot company, developers, and government, and proposed concrete pathways to achieve continuity. Conversely, those who opposed or were uncertain about it offered equally strong justifications.

\subsubsection{Fulfilling Continuous Service Through Technical, Financial, and External Safeguards (RQ2a)}
\label{sec:fulfilling_service}
\mbox{}\

\noindent When asked who should be responsible for the continuous service of the Moxie robot, participants identified the robot company, their software developers, and the government as key stakeholders.

\textbf{Implementing Technical Mechanisms for Post-Shutdown Independence.}
Participants argued that the primary responsibility for continuity lies with developers to design a robot not wholly dependent on its company. The most common suggestion was for developers to release a final ``legacy'' firmware update, as a participant stated:
\begin{quote}
\textit{Just the development team so that they can release a final firmware for the robot so that basic functions can be performed by it indefinitely.}
\end{quote}

This case places the responsibility on the technical team to create a functional end-of-life plan. Others further focused on the initial design, arguing there should be: \textit{``perhaps lack of a need for it to connect to a company's external server to continue to operate.''} This highlights an expectation for built-in resilience. A more decentralized approach involved empowering users, with a participant suggesting that company and software developers could: \textit{``open source the software to the customers so they can be able to run their own server for the robots, if they want to.''} This solution shifts power from the corporation to the family users for the robot product to exist long after the original company has dissolved.

\textbf{Establishing Sustainable Financial Models for Long-Term Operation.}
Participants placed much responsibility on the robot company to establish a financial model that ensures its own long-term survival. One recurring suggestion was a user-funded subscription model. A participant offered an example:
\begin{quote}
\textit{Customers could be charged an extra fee... a customer can pay 10 per month to keep the robot working and help the company from being bankrupt.}
\end{quote}

This creates a direct, sustainable revenue stream tied to the existing family user base. Participants also argued that the company and developers must secure strong financial backing from the outset, with one noting the importance of having the \textit{``backing of strong financial people that can invest in the company.''} 

\textbf{Leveraging External Governance for Corporate Guarantees.}
When a company's planning fails, participants proposed that external entities or pre-established guarantees should serve as a fail-safe. This included roles for third-party regulators, the government, and the robot company itself. Some suggested a formal regulatory layer, stating, \textit{``There should be another third-party to regulate this issue.''} Direct government intervention was also seen as a viable solution, with one participant stating: \textit{``Well, I feel the government should be able to buy the company to keep it running.''} This reflects a belief that children's emotional well-being is a public interest that may warrant state and government intervention.

\subsubsection{Opposing a Continuous Service due to Business Realities and Developmental Concerns (RQ2b)}
\label{sec:opposing_service}
\mbox{}\

\textbf{Asserting the Practical Impossibility of Continuous Service.}
The most common reason why participants opposed or were uncertain about a continuous service was grounded in business realities. Participants argued that forcing a failed company to maintain services is neither financially feasible nor legally obligated. They frequently questioned the logistics: \textit{``how is that running if the company behind it is bankrupt? Who is funding the operation of this network?''} Another participant stated: \textit{``If a company can't afford to stay open, there's nothing they can do about that.''} This highlights the view that emotional attachment cannot override economic constraints. 

Participants also argued that a company’s obligations are limited by market principles. One explained: \textit{``This is a business. a for-profit business. They have an obligation to their shareholders and were unable to fulfill that obligation.''} This viewpoint defines the company’s primary duty as being to its financial stakeholders, not a perpetual service guarantee. Another echoed this, stating, \textit{``Nothing lasts forever, and the company is a business first.''}

\begin{figure*}[h] 
    \centering
    \includegraphics[width=1\textwidth]{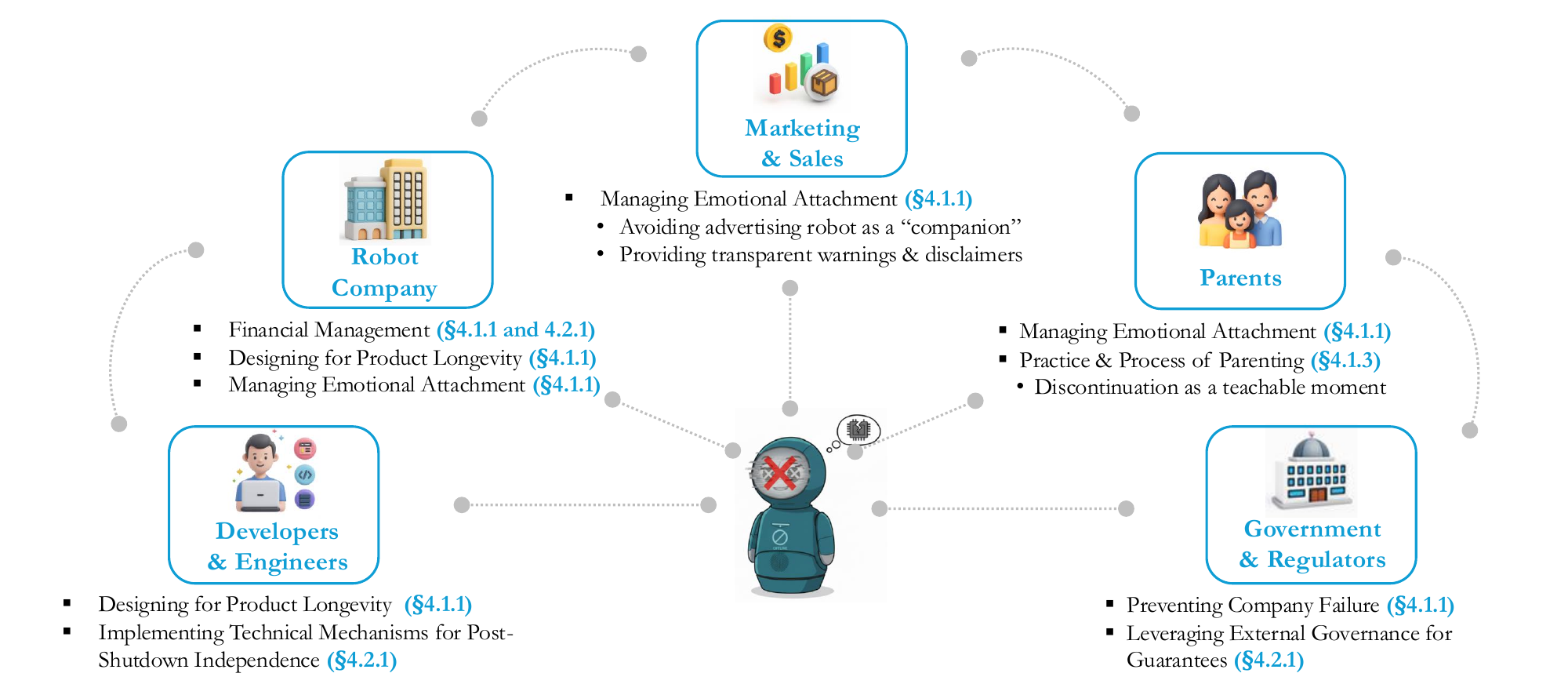}
    
    \caption{A shared responsibility framework for mitigating the emotional risks in child-robot companionship. This framework illustrates the distribution of responsibility among key stakeholders, including the Robot Company, Developers, Marketing, Parents, and Government.}
    \label{fig:shared_responsibility_framework}
\end{figure*}

\textbf{Framing Continuous Service as a Detriment to Healthy Child Development.}
A second group opposed the requirement because they believed continued reliance on the robot was itself harmful. They framed the deep emotional relationship with the Moxie robot as an unhealthy dependency. One participant commented: \textit{``because it can become an addiction of sorts and a very unhealthy attachment.''} Participants further argued that the robot actively hinders the development of social skills by replacing authentic human connections. As one participant put it: \textit{``I think I child should be interacting with actual humans rather than a robot. I feel like it would be unhelpful to a child to continuously use the robot.''} Another participant also echoed this view, stating that \textit{``children need socialization with other human children more than a robot.''}

\textbf{Accepting Robot Discontinuation as a Necessary Life Lesson.}
Finally, some participants framed the Moxie robot’s shutdown as a life lesson in loss and resilience. From this perspective, shielding a child from it would be a disservice to their long-term emotional growth. A participant stated: \textit{``Sometimes it just is not possible, and everyone needs to confront losses in life.''} Another reinforced this idea, noting: \textit{``A child can become attached to anything that could ultimately break or stop working.''} This suggests that learning to cope with emotional disappointment is a part of growing up.

\section{DISCUSSION}

\subsection{Theoretical Implications: A Shared Responsibility Framework} 


Our findings extend shared responsibility frameworks from healthcare ~\cite{elwyn2012shared, shay2015evidence} and AI ethics ~\cite{arroyo2022shared, muhlhoff2025collective} to the context of child-robot companionship, culminating in a shared responsibility framework (Figure~\ref{fig:shared_responsibility_framework}). The framework first positions corporate stakeholders, including the company, developers, and marketers, as the architects of children's emotional bond. Our participants overwhelmingly assigned foundational responsibility to the robot company for its failed financial plan and to developers for designing a product without end-of-life contingencies, such as offline functionality (§\ref{sec:shared_framework}, §\ref{sec:fulfilling_service}). This validates arguments that AI harm often emerges from a chain of decisions rather than a single point of failure ~\cite{arroyo2022shared}. While existing literature explains how these bonds form through mechanisms like the CASA paradigm ~\cite{nass1994computers} and anthropomorphism ~\cite{epley2007seeing}, our study articulates the public’s attribution of responsibility when these mechanisms are emotionally severed. Likewise, participants’ condemnation of marketing that personified the Moxie robot as a perpetual ``friend'' (§\ref{sec:shared_framework}) confirms warnings that such narratives exploit users' innate ``sociality motivation'' ~\cite{epley2007seeing} and create unsustainable expectations of permanence ~\cite{gunawan2025promises}.

Our framework then delineates the interdependent roles of parents and government as guardians, revealing a tension between individual and systemic duties. Our findings show a strong public expectation for parents to act as the ultimate gatekeepers by managing emotional attachment and teaching resilience (§\ref{sec:shared_framework}), a view that supports calls in prior work for greater parental AI literacy ~\cite{rosman2025exploring, ho2025empowering}. However, our study complicates this by uncovering a stark divide between parents and non-parents (§\ref{sec:parental_divide}); where non-parents critiqued the foundational choice to purchase the robot, parents focused on the practical difficulties of mediating the relationship, exposing a gap between societal expectation and the lived reality of raising children with technology. This gap reinforces the need for external oversight, a role participants assigned to the government (§\ref{sec:fulfilling_service}). This public demand for regulatory action empirically confirms the ``regulatory vacuum'' identified in prior work ~\cite{de2025unregulated}, while our finding of a sharp political divide (§\ref{sec:political_divide}) on the nature of this intervention, systemic versus market-based, shows a practical barrier to implementing the comprehensive frameworks others have proposed ~\cite{mclennan2014exposing}.

Ultimately, our framework illustrates that the emotional harm from Moxie robot discontinuation is not a singular failure but an emergent property of an interconnected system, where the actions and inactions of each stakeholder compound on the others. For example, marketing promises of a perpetual ``friend'' (§\ref{sec:shared_framework}) intensify the burden on parents to set boundaries (§\ref{sec:parental_divide}), while developers' failure to engineer for product longevity (§\ref{sec:fulfilling_service}) creates a crisis that parents are ill-equipped to handle, prompting calls for government intervention (§\ref{sec:political_divide}). By detailing these interdependencies, our work moves beyond simply identifying that responsibility is shared. Instead, we offer empirically-derived duties specific to child-robot companionship, detailing how this responsibility is distributed, contested, and co-constructed in practice.

\subsection{Design Implications}
\mbox{}\

\textbf{Mandatory Offline Functionality} (§\ref{sec:shared_framework}): Participants in their responses repeatedly identified server dependency as a preventable technical failure. This finding mandates that companion robots include core offline capabilities—basic conversation, stored memories, and play functions—that persist independent of company infrastructure.

\textbf{Layered Robotic Architecture} (§\ref{sec:fulfilling_service}): Related to the mandating offline functionality, our findings also revealed frustration with all-or-nothing functionality, with participants proposing firmware or pre-recorded feedback to keep a robot companion alive. This suggests implementing a layered structure when designing robot companions: full functionality when connected to the Internet, reduced local network mode, and basic offline companion features.

\textbf{Transparent Attachment Indicators} (§\ref{sec:shared_framework}): Given the fact that most participants identified robot companies and parent stakeholders, there is a need for transparent communication mechanisms between these stakeholders. Companion robots should include built-in features that make the nature and intensity of child-robot relationships visible to parents: alerts clarifying the robot's artificial nature at regular intervals, dashboards displaying interaction frequency patterns, analysis of attachment language used by the child, and dependency indicators that flag concerning usage patterns.

\subsection{Policy Implications}
\mbox{}\

\textbf{Emotional AI Risk Classification}: Participants' emphasis on the unique nature of emotional attachment of children to robots (§\ref{sec:opposing_service}) necessitates comprehensive risk assessment across the spectrum of AI companions. Different combinations of technology type (text-based chatbots, voice assistants, embodied robots) and user demographics (children, elderly, individuals with disabilities) create distinct risk profiles requiring tailored safeguards. This graduated classification system would enable proportional regulation that maximizes available resources to protect vulnerable populations without unnecessarily stifling technological innovations.

\textbf{Pre-Market Sustainability Evaluation}: The polarized views on service continuity as revealed in this study (§\ref{sec:political_divide}, §\ref{sec:parental_divide}, and §\ref{Sec:RQ2}) reflect tension between participants recognizing the business volatility in developing cutting-edge technologies, and the devastating emotional damages to users due to business failure. Rather than perpetual service mandates, regulation could require companies to do a comprehensive pre-market continuity planning through insurance, escrow funds, or documented transition protocols.

\textbf{Mandatory User Education} (§\ref{sec:fulfilling_service}): The gap between participants' desire for open-source empowerment and the OpenMoxie reality, where technical barriers excluded most families, highlights the need for accessible user education. Companies must provide non-technical explanations of robot dependency, plain-language maintenance guides, clear discontinuation warnings, and community support connections at purchase. This shifts from relying on highly specialized knowledge to genuinely empowering families with actionable literacy and tools they can actually use.

\textbf{Marketing and Disclosure Standards}: Participants across political lines criticized marketing that one-sidedly framed Moxie as a companion constantly "be there" for kids (§\ref{sec:opposing_service}). Regulators like the FTC must prohibit advertising and marketing language implying permanence or unconditional availability and claims about emotional support without accompanying risk disclosure. Companies must be required to provide mandatory pre-purchase disclosures about attachment risks comparable to pharmaceutical side-effect warnings, and age-appropriate materials explaining the robot's mechanical nature to children.

\textbf{Tiered Regulatory Approach}: Last, the political ideological divide (§\ref{sec:political_divide}) revealed in this study, where Democrats advocated government intervention versus Republicans emphasizing market solutions, suggests that uniform federal mandates will likely face resistance, though consensus has been reached across the political spectrum regarding the deeply troubling case of Moxie shutdown. Policy makers can, instead, explore a tiered regulation approach with federal minimums and state-level additions to accommodate these ideological differences while ensuring baseline protections.

\section{LIMITATIONS \& FUTURE WORK}
Like all research, this study has limitations that suggest directions for future work. First, our sample consisted of U.S. adults viewing videos of children's reactions rather than directly surveying affected families. While this approach avoided potential harm to grieving children and provided diverse perspectives on responsibility attribution, it may not fully capture the lived experiences of families who invested financially and emotionally in Moxie. Future research should explore ethical methods for conducting longitudinal interviews with affected families after sufficient time has passed for emotional processing.

Second, the video stimulus captured immediate reactions to Moxie's shutdown announcement, but did not account for subsequent developments, particularly the OpenMoxie project, a community led initiative requiring significant programming skills to maintain basic functionality through local hosting and demanding programming skills. This temporal limitation could affect our understanding of ``right-to-use'' implementation. Future research should examine how the public views such high-barrier solutions and how to keep them easily accessible.

\section{CONCLUSIONS}
Through qualitative analysis of public responses to the Moxie robot shutdown, this study empirically examined shared responsibility attribution for children's emotional suffering following the Moxie robot shutdown, revealing complex stakeholder dynamics and ideological tensions around companion technology governance. Our findings demonstrate that the safe emotional bonds between children and robots demand shared responsibilities across multiple stakeholders, with attribution patterns significantly shaped by political ideology and parental status. As companion robots increasingly touch the critical developmental stages of children, our findings underscore the urgent need for dialogues on shared responsibilities and the development of comprehensive regulations that safeguard healthy child-robot companionship.



\bibliographystyle{ACM-Reference-Format}
\bibliography{main}

\appendix

\section{Participant Demographic Information}
\label{appendix_demographic}
This study included 72 participants who provided qualitative responses. Of these, 70 participants completed the demographic survey, the details of which are presented in Table~\ref{tab:demographics}. The participants had a mean age of 41.8 years (SD=11.87), and the sample was evenly split by sex (50.0\% male, 50.0\% female) and parental status (50.0\% with children, 50.0\% without). The most common educational attainment was a technical or undergraduate degree (51.4\%), and the largest political affiliation group was identified as `Independent/Other' (40.0\%).

\begin{table}[t]
\small
\caption{Participant demographics. While 72 participants provided qualitative responses, demographic data was collected from N=70.}
\label{tab:demographics}
\centering
\setlength{\tabcolsep}{6pt}
\begin{tabularx}{\columnwidth}{@{}p{0.3\columnwidth}>{\raggedright\arraybackslash}X>{\raggedleft\arraybackslash}r@{}}
\toprule
\textbf{Demographic} & \textbf{Category} & \textbf{N (\%)} \\
\midrule
Age (N=70) & Mean (SD) & 41.8 (11.87) \\
           & Range (Min, Max) & 21, 70 \\
\addlinespace
Sex (N=70) & Female & 35 (50.0\%) \\
           & Male & 35 (50.0\%) \\
\addlinespace
Education (N=70) & Community college/Undergraduate & 36 (51.4\%) \\
                 & High school/Secondary & 18 (25.7\%) \\
                 & Graduate/Doctorate & 16 (22.9\%) \\
\addlinespace
Political Affiliation \\ (N=70) & Independent/Other & 28 (40.0\%) \\
                             & Republican & 16 (22.9\%) \\
                             & Democrat & 14 (20.0\%) \\
                             & None & 12 (17.1\%) \\
\addlinespace
Has Children \\ (N=70) & Yes & 35 (50.0\%) \\
                    & No & 35 (50.0\%) \\
\bottomrule
\end{tabularx}
\end{table}

\clearpage
\onecolumn

\section{Survey Design}
\label{appendix_surveydesign}
The following table details the open-ended questions presented to participants in the qualitative survey. The survey was structured to first gather responses on responsibility attribution before probing opinions on the continuity of service.

\small
\begin{longtable}{p{0.95\textwidth}}
\toprule
\textbf{Survey Questions} \\
\midrule
\endfirsthead
\toprule
\textbf{Survey Questions (continued)} \\
\midrule
\endhead
\multicolumn{1}{c}{\textbf{Informed Consent - Responsibility Attribution}} \\
\midrule
Think about what happened to the children in the video, who you think bears responsibility for this situation? \\
$\square$ Robot manufacturers 
$\square$ Robot companies 
$\square$ Parents 
$\square$ Government or legislation 
$\square$ Software developers/engineers
$\square$ Marketing/sales teams
$\square$ Other (please type your answer) \\
\midrule

\multicolumn{1}{c}{\textbf{Robot Companies}} \\
\midrule
You identified robot companies as responsible for children's suffering when they lose their robot companions. Why? \\
What specific responsibilities should robot companies have to prevent children's suffering like this from happening again? \\
\midrule

\multicolumn{1}{c}{\textbf{Parents}} \\
\midrule
You identified parents as responsible for children's suffering when they lose their robot companions. Why? \\
What specific responsibilities should parents have to prevent children's suffering like this from happening again? \\
\midrule

\multicolumn{1}{c}{\textbf{Government Regulators}} \\
\midrule
You identified government or legislation as responsible for children's suffering when they lose their robot companions. Why? \\
What specific responsibilities should government or legislation have to prevent children's suffering like this from happening again? \\
\midrule

\multicolumn{1}{c}{\textbf{Software Developers/Engineers}} \\
\midrule
You identified software developers/engineers as responsible for children's suffering when they lose their robot companions. Why? \\
What specific responsibilities should software developers/engineers have to prevent children's suffering like this from happening again? \\
\midrule

\multicolumn{1}{c}{\textbf{Marketing/Sales Teams}} \\
\midrule
You identified marketing/sales teams as responsible for children's suffering when they lose their robot companions. Why? \\
What specific responsibilities should marketing/sales teams have to prevent children's suffering like this from happening again? \\
\midrule

\multicolumn{1}{c}{\textbf{PARTY I-V}} \\
\midrule
You identified [custom party] as responsible for children's suffering when they lose their robot companions. Why? \\
What specific responsibilities should [custom party] have to prevent children's suffering like this from happening again? \\
\midrule

You identified [custom party] as responsible for children's suffering when they lose their robot companions. Why? \\
What specific responsibilities should [custom party] have to prevent children's suffering like this from happening again? \\
\midrule

You identified [custom party] as responsible for children's suffering when they lose their robot companions. Why? \\
What specific responsibilities should [custom party] have to prevent children's suffering like this from happening again? \\
\midrule

You identified [custom party] as responsible for children's suffering when they lose their robot companions. Why? \\
What specific responsibilities should [custom party] have to prevent children's suffering like this from happening again? \\
\midrule

You identified [custom party] as responsible for children's suffering when they lose their robot companions. Why? \\
What specific responsibilities should [custom party] have to prevent children's suffering like this from happening again? \\
\midrule

\multicolumn{1}{c}{\textbf{Right to Use}} \\
\midrule
When a child has formed a strong emotional bond with a robot companion like Moxie, should the services be required to continue? \\
Why do you think that the services should be required to run continuously? \\
Why do you think that the services should NOT be required to run continuously? \\
Why are you unsure about if the services should be required to run continuously? \\
Who (e.g., robot company, government, etc.) specifically should be responsible for keeping the services running continuously, and what exactly should they be required to do? \\
What systems, resources, or mechanisms would need to be in place to maintain continuous services of a robot companion? \\

\bottomrule
\end{longtable}

\end{document}